\pdfoutput=1

\documentclass[11pt]{article}

\usepackage{ACL2023}

\usepackage{times}
\usepackage{latexsym}
\usepackage{booktabs}
\usepackage{amsmath}
\usepackage{graphicx} 
\usepackage{multirow}
\usepackage{lipsum} 
\usepackage{float} 
\usepackage{float}       
\usepackage{booktabs}    
\usepackage{graphicx}    
\usepackage{caption}     
\usepackage{placeins} 

\usepackage{amsmath}
\usepackage{amssymb}
\usepackage{adjustbox}

\usepackage[T1]{fontenc}

\usepackage[utf8]{inputenc}

\usepackage{microtype}

\usepackage{inconsolata}

\title{\texttt{fact check AI} at SemEval-2025 Task 7: Multilingual and Crosslingual Fact-checked Claim Retrieval}

\author{Pranshu Rastogi \\
  Independent Researcher \\
  \texttt{rastogipranshu29@gmail.com}
}

\begin{document}
\maketitle
\begin{abstract}
SemEval-2025 Task 7: Multilingual and Crosslingual Fact-Checked Claim Retrieval is approached as a Learning-to-Rank task using a bi-encoder model fine-tuned from a pre-trained transformer optimized for sentence similarity. Training used both the source languages and their English translations for multilingual retrieval and only English translations for cross-lingual retrieval. Using lightweight models with fewer than 500M parameters and training on Kaggle T4 GPUs, the method achieved \textbf{92\%} \textbf{Success@10} in multilingual and \textbf{80\%} Success@10 in \textbf{5th} in \textbf{crosslingual} and \textbf{10th} in \textbf{multilingual} tracks. \\
Github-\href{https://github.com/pranshurastogi29/SemEval-2025-ACL-Multi-and-Crosslingual-Retrieval-using-Bi-encoders}{SemEval-2025-ACL-Multi-and-Crosslingual-Retrieval-using-Bi-encoders}
\end{abstract}

\section{Introduction}
The rapid spread and multilingual nature of online disinformation pose a significant challenge to traditional fact-checking workflows. \textit{SemEval-2025 Task 7: Multilingual and Crosslingual Fact-Checked Claim Retrieval} \cite{semeval2025task7} addresses this issue by aiming to automate the retrieval of previously verified claims across languages. The corpus for this task consists of fact-checked claims and social media posts in eight languages: French, Spanish, English, Portuguese, Thai, German, Malay, and Arabic. To further evaluate the generalization of the model, the evaluation step introduced two new unseen languages, Pol and Tur, underlining the demand for strong multilingual and cross-lingual retrieval approaches. Our approach tries to tackle these challenges by improving performance both in multilingual and crosslingual scenarios. 
We utilize lightweight, scalable transformer models that can be trained efficiently even on modest hardware. Our approach produces rich semantic representations and utilizes methods such as layer freezing and gradient checkpointing to tune the trade-off between efficiency and effectiveness. The outcome is a practical, adaptable tool for global misinformation detection that addresses both retrieval accuracy and computational scalability. We trained independent models for each language and utilized English translations of the source language inputs to boost the retrieval. For cross-lingual tasks, we predominantly used English translations to achieve consistency. 
In multilingual environments, we retrieved the top 10 fact verify claims from several models and ranked them according to their scores. For crosslanguage retrievals, we repeated the same process using outputs from a five-fold ensemble. This allowed us to combine diversity, accuracy, and speed very well. Our approach finally ended up being 5th in crosslanguage retrieval and 10th in multilingual retrieval. All models were less than 500M parameters, and training was
done using only two Kaggle T4 GPUs showing that high performance is attainable even with
constrained computational resources.

\section{Background}
The SemEval-2025 Task 7 is all about creating systems that can find reliable fact-checked claims for social media posts. This helps fact-checkers who deal with different languages. It’s tough to look for fact-checks in many languages by hand. So, we need to automate this to save time and effort.

There are two main parts to the task:\\
\textbf{Multilingual Retrieval}: Here, both the social media post and the fact-check are in the same language.\\
\textbf{Crosslingual Retrieval}: In this case, the post and the fact-check are in different languages. This needs strong techniques for matching across languages.

\subsection{Input and Output Format}
The input includes social media posts and fact-checks. Each has text, metadata, and English translations. The system gives a ranked list of up to 10 fact-checks for each post.

\textbf{Example Output:}
\small
\texttt{\{
"Post-12345": [987, 654, 321, 789, 456, 222, 111, 333, 555, 777]
\}}
\normalsize
\\
This means that Post-12345 is best matched with Fact-check IDs 987, 654, etc., sorted by relevance.

\subsection{Dataset Details}

The training dataset includes Fact-checks and Posts in 8 languages:
Arabic (ara), German (deu), English (eng), French (fra), Malay (msa), Portuguese (por), Spanish (spa), Thai (tha)
However, in the final test set, two surprise languages—\textbf{Tur (tur) and Pol (pol)}—were introduced, making the task more challenging as models needed to generalize to unseen languages without direct training data. As shown in Table~\ref{tab:train_test_data}

\begin{table*}[t]
    \centering
    \small 
    \renewcommand{\arraystretch}{1.2} 
    \setlength{\tabcolsep}{4pt} 

    \caption{Number of Post and Fact-check IDs in Train and Test Sets for Each Language}
    \label{tab:train_test_data}

    \begin{adjustbox}{width=\textwidth} 
    \begin{tabular}{|l|r|r|r|r|}
        \hline
        \multirow{2}{*}{\textbf{Multi-Lingual}} & \multicolumn{2}{c|}{\textbf{Train}} & \multicolumn{2}{c|}{\textbf{Test}} \\
        \cline{2-5}
        & \textbf{Post IDs} & \textbf{Fact-check IDs} & \textbf{Post IDs} & \textbf{Fact-check IDs} \\
        \hline
        Arabic (ara) & 14,201 & 676 & 500 & 21,153 \\
        German (deu) & 4,996 & 667 & 500 & 7,485 \\
        English (eng) & 85,734 & 4,351 & 500 & 145,287 \\
        French (fra) & 4,355 & 1,596 & 500 & 6,316 \\
        Malay (msa) & 8,424 & 1,062 & 93 & 686 \\
        Portuguese (por) & 21,569 & 2,571 & 500 & 32,598 \\
        Spanish (spa) & 14,082 & 5,628 & 500 & 25,440 \\
        Thai (tha) & 382 & 465 & 183 & 583 \\
        Pol (pol) & - & - & 500 & 8,796 \\
        Tur (tur) & - & - & 500 & 12,536 \\
        \hline
        \textbf{Cross-Lingual} & 153,743 & 4,972 & 8,276 & 272,256 \\
        \hline
    \end{tabular}
    \end{adjustbox}

\end{table*}

\subsection{Dataset Files}
The dataset has three key files:
\cite{pikuliak-etal-2023-multilingual}\\
1. \textbf{Fact-checks.csv} - This file has fact-check claims, titles, and URLs. It's available in both the original language and in English.

2. \textbf{Posts.csv} - Here, you'll find social media posts. It includes text from those posts, fact-checks taken from images, Meta’s verdicts like False Information, and their English translations.

3. \textbf{Fact-check-Post-mapping.csv} - This file connects social media posts to their fact-checks. It also shows the language pairs, like spa-eng, which means a Spanish post was fact-checked in English.

\subsection{Evaluation Metrics}
To assess system performance, we use:
\textbf{Success@10 (S@10)} – Measures whether at least one correct Fact-check appears in the top 10 retrieved results.

\section{System Overview}
Our system uses a Bi-encoder setup \cite{reimers2019sentencebertsentenceembeddingsusing}. This helps match Posts and Fact-checks quickly. We also use smart pooling methods to improve pre-trained Sentence similarity models. The model has a few important parts:

\subsection{Bi-Encoder Retrieval Model}
We use a Bi-encoder design. This means we have a \textbf{Post Encoder} and a \textbf{Fact-check Encoder}. They take inputs and turn them into dense vector representations. We use a pretrained transformer backbone for this. In each batch, we have posts and fact-checked claims. We then find the similarity between their embeddings with a simple math function. For training, we apply MNR Loss \cite{henderson2017efficientnaturallanguageresponse} and Vanilla Cross-Entropy Loss.

\subsection{Pretrained Transformer Encoder}
The encoders are initialized with publicly available transformer weights. Given an input sequence \( X = (x_1, x_2, ..., x_n) \), the model outputs contextualized token embeddings:

\begin{equation}
H = \mathcal{M}(X) \in \mathbb{R}^{n \times d}
\end{equation}

where \( H \) represents the hidden states, \( n \) is the sequence length, and \( d \) is the hidden dimension.

\subsection{Pooling Mechanisms}
To obtain a fixed-size sentence representation, we explore multiple pooling strategies:

\paragraph{Mean Pooling:} Computes the mean of token embeddings weighted by attention masks:
\begin{equation}
\text{MeanPooling}(H, A) = \frac{\sum_{i=1}^{n} H_i \cdot A_i}{\sum_{i=1}^{n} A_i + \epsilon}
\end{equation}
\paragraph{Attention Pooling with BiLSTM:} A bidirectional LSTM \cite{6795963} enhances contextual aggregation:
\begin{equation}
L = \text{BiLSTM}(H) \in \mathbb{R}^{n \times 2h}
\end{equation}
An attention mechanism assigns dynamic weights to tokens:
\begin{equation}
\alpha = \text{softmax}(W_a L), \quad S = \sum_{i=1}^{n} \alpha_i L_i
\end{equation}
where \( W_a \) is a learnable dense layer.

\subsection{Similarity Function}
Relevance between a Post embedding \( q \) and a Fact-check embedding \( c \) is computed using \textbf{temperature-scaled cosine similarity}:

\begin{equation}
S(q, c) = \frac{\text{cos}(q, c)}{T}
\end{equation}

where \( T \) is a temperature parameter kept at 0.05.

\subsection{Contrastive Learning with MNR Loss \cite{henderson2017efficientnaturallanguageresponse} and Cross Entropy Loss}
To optimize retrieval, we employ a \textbf{MNR Loss} \cite{henderson2017efficientnaturallanguageresponse} function that maximizes similarity for positive pairs:

\begin{equation}
\mathcal{L} = - \sum_i \log \frac{\exp(S_{ii})}{\sum_j \exp(S_{ij})}
\end{equation}

To improve efficiency, we use a symmetric formulation:

\begin{equation}
\mathcal{L} = \frac{1}{2} \left( \text{CrossEntropy}(S, y) + \text{CrossEntropy}(S^T, y) \right)
\end{equation}

\subsection{End-to-End Flow}\label{subsec:e2e}
Posts and Fact-checked claims are processed through separate encoders that shares weights. Get embeddings using LSTM \cite{6795963} or by Mean Pooling. Similarities are computed via temperature-scaled cosine similarity, The model is optimized using Loss functions defiend above to rank relevant claims higher. 

In this architecture defined above I have initialised Encoder models with these Pre-trained Models
\begin{enumerate}
\item The \href{https://huggingface.co/NovaSearch/stella_en_400M_v5}{NovaSearch/stella-en-400M-v5} model \cite{zhang2025jasperstelladistillationsota}, This model is ranked 41st on the \href{https://huggingface.co/spaces/mteb/leaderboard}{MTEB leaderboard}, has 435M parameters and balances scalability with retrieval accuracy.  

\item The \href{https://huggingface.co/intfloat/multilingual-e5-large-instruct}{intfloat/multilingual-e5-large-instruct} model \cite{wang2024multilingual}, with 24 layers and 1024 embedding size, ranks 22nd on the \href{https://huggingface.co/spaces/mteb/leaderboard}{MTEB leaderboard} with 560M parameters. Built on xlm-roberta-large \cite{conneau2020unsupervisedcrosslingualrepresentationlearning}, it supports 100 languages. This model was the primary choice for direct training on source languages.
\item The \href{https://huggingface.co/mixedbread-ai/mxbai-embed-large-v1}{mixedbread-ai/mxbai-embed-large-v1} model \cite{emb2024mxbai} is a state-of-the-art English embedding model that balances efficiency and performance with 335M parameters, currently ranking 49th on the \href{https://huggingface.co/spaces/mteb/leaderboard}{MTEB} leaderboard.
\end{enumerate}

All the models we selected have fewer than 500M parameters, making them suitable for training on free online GPUs. This allows for extended training, which helps produce richer deep text representations. Our training architecture is designed to balance computational efficiency and retrieval accuracy. We achieve this by using lightweight transformer models with techniques like gradient checkpointing and selective layer freezing, ensuring scalability without compromising performance.

During evaluation, we generate embeddings for both social media posts and fact-checked claims. To retrieve matches, we employ semantic search, which compares the embeddings to identify the most similar pairs. Our system uses an optimized retrieval pipeline that indexes the data efficiently, enabling fast similarity matching. We then select the top 10 most relevant claims for each query post.

We also experimented with other models like 
\href{https://huggingface.co/jinaai/jina-embeddings-v3}{\texttt{jina-embeddings-v3}} and \href{https://huggingface.co/HIT-TMG/KaLM-embedding-multilingual-mini-v1}{\texttt{KaLM-embedding-multilingual-mini-v1}}, which are currently ranked 20th and 19th on the MTEB leaderboard. While both models performed reasonably well with average scores of 58.37 (Jina) and 57.05 (KaLM)—they fell short of the performance achieved by \textbf{multilingual-e5-large-instruct}, which scored 63.23 on average. These comparisons informed our decision to prioritize higher-performing models for downstream tasks.

Before training and evaluation, we applied preprocessing steps to improve data quality. This included filtering out short or symbol-heavy text, removing URLs, emojis, and excessive whitespace, and standardizing punctuation. For OCR-extracted data, we excluded noisy text segments. Twitter-specific cleaning involved replacing image references and shortened URLs with placeholders to maintain consistency across samples.

\begin{table*}
    \centering
    \renewcommand{\arraystretch}{1.2} 
    \resizebox{\textwidth}{!}{  
    \begin{tabular}{l|c|c|c|c|c|c|c|c|c|c|c|c}
        \toprule
        \textbf{Model} & \textbf{Cross-Lingual (avg)} & \textbf{Multilingual (avg)} & \textbf{eng} & \textbf{fra} & \textbf{deu} & \textbf{por} & \textbf{spa} & \textbf{tha} & \textbf{msa} & \textbf{ara} & \textbf{tur} & \textbf{pol} \\
        \midrule
        multilingual-e5-large-instruct (Best Fold) & 0.7685 & 0.9091 & 0.864 & 0.932 & 0.904 & 0.85 & 0.924 & 0.967 & 1.0 & 0.942 & 0.864 & 0.848 \\
        stella-en-400M-v5 (Best Fold) & 0.76 & 0.9095 & 0.852 & 0.913 & 0.902 & 0.84 & 0.917 & 0.978 & 1.0 & 0.931 & 0.848 & 0.874 \\
        mxbai-embed-large-v1 (Best Fold) & 0.7275 & 0.8972 & 0.844 & 0.92 & 0.884 & 0.828 & 0.902 & 0.978 & 0.989 & 0.932 & 0.844 & 0.852 \\
        multilingual-e5-large-instruct (Ensemble) & 0.7975 & 0.9232 & 0.882 & 0.944 & 0.926 & 0.866 & 0.942 & 0.995 & 0.989 & 0.940 & 0.884 & 0.864 \\
        stella-en-400M-v5 (Ensemble) & 0.7638 & 0.9140 & 0.876 & 0.936 & 0.906 & 0.870 & 0.946 & 0.978 & 1.000 & 0.932 & 0.844 & 0.852 \\
        mxbai-embed-large-v1 (Ensemble) & 0.7712 & 0.9146 & 0.852 & 0.934 & 0.892 & 0.868 & 0.930 & 0.989 & 0.978 & 0.956 & 0.886 & 0.860 \\
        \bottomrule
    \end{tabular}
    }
    \caption{Cross-Lingual and Multilingual Success@10 Breakdown on Leaderboard}
    \label{tab:crosslingual_mono_combined}
\end{table*}
\section{Experimental Setup}

\subsection{Multi-Lingual Training}
For Multilingual bi-encoder \cite{reimers2019sentencebertsentenceembeddingsusing} training, the full dataset was utilized for each language, benefiting low-resource languages like German, French, Malay, and Thai Table \ref{tab:train_test_data}. Without synthetic or external data, multiple training epochs allowed the model to capture both semantic and syntactic patterns more effectively. 

\subsubsection{Evaluation Measures}
The model is trained with contrastive loss(MNR loss \cite{henderson2017efficientnaturallanguageresponse} and cross-entropy over similarity scores) to optimise retrieval performance. At training time, similarity scores between
posts and fact-checked embeddings are calculated, which ensures correct matches rank higher. The
performance of the model is measured on Success@10 metrics

\subsubsection{Training Strategy}
For training, we used the Full dataset for all source languages. We did multiple epochs of training to improve learning. When training with English translations, we picked a random \textbf{30\%} sample of negative cases for testing. We used a random seed of \textbf{42}.

\subsubsection{Backbone Model}
\texttt{multilingual-e5-large-instruct} \cite{wang2024multilingual} and \texttt{stella-en-400M-v5} \cite{zhang2025jasperstelladistillationsota} is fine-tuned with gradient checkpointing disabled

\subsubsection{Pooling Mechanisms}
\begin{itemize}
    \item \textbf{Mean Pooling}: Mean of hidden states unless "cls" token-based pooling is specified.
    \item \textbf{Attention Pooling}: Uses a bidirectional LSTM followed by an attention mechanism to weight token embeddings dynamically.
\end{itemize}

\begin{table}[h]
    \centering
    \begin{tabular}{ll}
        \toprule
        \textbf{Parameter} & \textbf{Value} \\
        \midrule
        Batch Size & 24 \\
        Number of Epochs & 20 \\
        Warmup Steps & 400 \\
        Mixed Precision Training & \texttt{float16} \\
        \bottomrule
    \end{tabular}
    \caption{MultiLingual Training Configuration}
    \label{tab:multilingual-config}
\end{table}

\subsubsection{Optimizer}
AdamW optimizer is configured with Learning Rate: $5 \times 10^{-6}$ for transformer, $1 \times 10^{-4}$ for custom layers. Weight Decay: 0.005. Gradient Clipping: Clip Value = 1.0

\subsection{Cross-Lingual Training}
For cross-lingual training, we re-used the same bi-encoder models but trained them on English-translated text by default. Rather than training on the entire dataset, we split the training data into 5 folds. This enabled us to train several models on varying data distributions, which promoted robustness. The text diversity mitigated overfitting and improved generalization, especially in low-resource language situations.

\subsubsection{Training Strategy}\label{subsec:train_strategy}
We used English-translated training data to enhance training data and limit overfitting. This method also improved performance in low-resource environments. Instead of synthetic data, we observed that ensembling English-trained models was a more efficient method for cross-lingual generalisation. This process was further amplified using 5-fold ensembling, which showed consistent improvement in retrieval performance. As indicated by Table \ref{tab:crosslingual_mono_combined}, ensembling resulted in significant improvements in all models. For instance, \textbf{multilingual-e5-large-instruct} went from \textbf{0.7685} to \textbf{0.7975} in cross-lingual retrieval, and from \textbf{0.9091} to \textbf{0.9232} in multilingual retrieval.

\subsubsection{Backbone Model}
\texttt{multilingual-e5-large-instruct} \cite{wang2024multilingual} and \texttt{stella-en-400M-v5} \cite{zhang2025jasperstelladistillationsota} and \texttt{mxbai-embed-large-v1} \cite{emb2024mxbai} were fine-tuned with gradient checkpointing disabled.

\subsubsection{Pooling Mechanisms}
The Pooling Mechanisms was similar to that in Multi-Lingual retraining

\begin{table}[h]
    \centering
    \begin{tabular}{ll}
        \toprule
        \textbf{Parameter} & \textbf{Value} \\
        \midrule
        Batch Size & 36 \\
        Number of Epochs & 10 \\
        Warmup Steps & 500 \\
        Mixed Precision Training & \texttt{float16} \\
        \bottomrule
    \end{tabular}
    \caption{Cross Lingual Training Configuration}
    \label{tab:training-config}
\end{table}

Optimizer configuration was similar to that in Multilingual settings

\section{Results}

\textbf{Multi-Lingual Rank} \textbf{10th} (S@10 avg = 0.9232)

\subsection*{Performances}
\begin{table}[h]
    \centering
    \caption{Performance of \texttt{fact check AI} ccross Languages}
    \begin{tabular}{lcc}
        \toprule
        \textbf{Metric} & \textbf{Score} & \textbf{Rank} \\
        \midrule
        S@10 (avg)      & 0.923178 & 10.0 \\
        S@10 (eng)      & 0.882    & 7.0  \\
        S@10 (fra)      & 0.944    & 5.0  \\
        S@10 (deu)      & 0.926    & 6.0  \\
        S@10 (por)      & 0.866    & 9.0  \\
        S@10 (spa)      & 0.942    & 7.0  \\
        S@10 (tha)      & 0.994536 & 4.0  \\
        S@10 (msa)      & 0.989247 & 8.0  \\
        S@10 (ara)      & 0.94     & 9.0  \\
        S@10 (tur)      & 0.884    & 10.0 \\
        S@10 (pol)      & 0.864    & 10.0 \\
        \bottomrule
    \end{tabular}
\end{table}

\subsection*{Insights on \texttt{fact check AI} Rankings and Scores}

\textbf{Overall Performance}:\\
Ranked \textbf{10th overall} with an average S@10 score of \textbf{0.923178}.While the model performs decently across languages, it falls behind top teams in consistency.

\textbf{Strongest Languages}:\\
Thai (S@10 = 0.994536, Rank = 4th) and French (S@10 = 0.944, Rank = 5th) are its best-performing languages. The model competes well in these languages, staying in the upper half of rankings.

\textbf{Weakest Languages}:\\
Pol (S@10 = 0.864, Rank = 10th) and Tur (S@10 = 0.884, Rank = 10th) have the lowest rankings.

\textbf{English Performance (Rank = 7th)}:\\
Middle of the pack, meaning the model isn’t optimized exclusively for English but is balanced across multiple languages.

Generalization Strength vs. Overfitting Risks: The model’s strong performance in under-represented languages like Thai and Malay suggests good generalization capabilities. In contrast, lower scores in languages such as English and Pol may point to potential domain overfitting or dataset imbalance. The wide performance range (S@10 from 0.864 to 0.9945) underscores the need for better multilingual adaptation, particularly for morphologically rich languages like Pol and Tur.

\section{Cross-Lingual Performance Analysis of \texttt{fact check AI}}

\subsection{Comparison with Top Teams}
\begin{table}[h]
    \centering
    \begin{tabular}{clc}
        \toprule
        Rank & Team Name & S@10 (avg) \\
        \midrule
        1 & PINGAN AI & 0.85875 \\
        2 & PALI & 0.82675 \\
        3 & Sherlock & 0.8245 \\
        4 & TIFIN India & 0.81025 \\
        \textbf{5} & \textbf{\texttt{fact check AI}} & \textbf{0.7975} \\
        \bottomrule
    \end{tabular}
    \caption{Cross-lingual ranking of \texttt{fact check AI} in comparison to other teams.}
    \label{tab:ranking}
\end{table}

\subsection{Overall Performance}
Our Cross lingual model ranks \textbf{5th}, achieving an average S@10 score of \textbf{0.7975}. It performs competitively in the upper half but remains behind the top-performing teams.

\subsection{Risks and Performance Insights}\label{subsec:preformance}
A key challenge is the potential for fact-check retrieval error, especially for low-resource or morphologically complex languages where semantic representations are less stable. The variation in performance across languages (S@10 from 0.864 to 0.9945) indicates probable training biases or domain overfitting. Examination showed solid performance where claims and posts had evident semantic overlap, even for under-represented languages like Thai and Malay. Performance dropped in morphologically dense languages like Pol and Tur, where inflectional complexity concealed
semantic similarity. Cross-lingual retrieval was particularly prone to mistakes with regard to idiomatic expressions or culturally specific references, revealing the difficulty in modeling deeper semantic subtleties between languages. Also, incorrect fact-checks were sometimes strongly ranked because of surface similarities e.g., common named entities—despite factual variations. These issues identify the risk of false positives where language-specific context is not correctly addressed. To mitigate these threats, improved multilingual adaptation by means of methods like language-aware fine-tuning, balanced sampling, and dynamic ensembling can help close performance disparity. Performance Risks and Insights, 

\section{Acknowledgments}
We would like to thank the SemEval-2025 Task 7 organizers for providing the cross-lingual and multilingual fact-check retrieval corpus and evaluation setup.We thank the anonymous reviewers very much for their thoughtful and constructive comments, which helped significantly in improving the quality of our manuscript. Relevant answers of your reviews can be found in the sections (\nameref{subsec:train_strategy})~\ref{subsec:train_strategy}, (\nameref{subsec:e2e}) ~\ref{subsec:e2e} and (\nameref{subsec:preformance}) ~\ref{subsec:preformance} in our paper.

\section{Conclusion}
We evaluated how effectively different multilingual em-embedding models perform for Multi and cross-language retrieval. We established that their effectiveness differs with the language, especially in low-resource environments like Pol and Tur Table \ref{tab:crosslingual_mono_combined}. Combining methods helped in improving retrieval performance, but variances still existed. Multilingual-e5-large-instruct worked best in single-language settings and cross-language situations. On the other hand, stella-en400M-v5 and mxbai-embed-large-v1 did not improve single-language performance. Going forward, we need to improve cross-language re-trievals more stable. We can achieve this by incorporating more data for less resource-full languages, tuning to specialized domains, and conbining various modeling strategies. This will enable real-world fact-check retrieval systems to become more robust.

\bibliography{fact_check_AI_at_semeval25_Task_7__Multilingual_and_Crosslingual_Fact_Checked_Claim_Retrieval}

\bibliographystyle{acl_natbib}

\end{document}